\documentclass[letterpaper, 10 pt, journal, twoside]{ieeetran}

\IEEEoverridecommandlockouts
\usepackage{cite}
\usepackage{amsmath,amssymb,amsfonts}
\DeclareMathOperator*{\argmax}{arg\,max}
\DeclareMathOperator*{\argmin}{arg\,min}
\usepackage[ruled,vlined]{algorithm2e}
\usepackage{graphicx}
\usepackage{textcomp}
\usepackage{siunitx}
\usepackage{xfrac}
\usepackage{makecell}
\usepackage{comment}
\usepackage{caption}
\usepackage{subcaption}

\usepackage{xcolor}
\usepackage{hyperref}
\usepackage{booktabs}
\usepackage{multirow}
\let\labelindent\relax
\usepackage[inline]{enumitem}
\newcommand*{\minOp}{\operatornamewithlimits{min}\limits}

\newcommand{\cmt}[1]{{\color{black}{#1}}} 
\newcommand{\cmtb}[1]{{\color{black}{#1}}} 

\begin{document}

\title{
Advanced Manufacturing Configuration\\by Sample-efficient Batch Bayesian Optimization
}

\author{
Xavier Guidetti$^{1,2}$,
Alisa Rupenyan$^{1,2}$, 
Lutz Fassl$^{3}$,
Majid Nabavi$^{3}$, and
John Lygeros$^{1}$

\thanks{Manuscript received: April 29, 2022; Revised August 8, 2022; Accepted September 11, 2022.}
\thanks{This paper was recommended for publication by Editor Ashis Banerjee upon evaluation of the Associate Editor and Reviewers' comments.}
\thanks{This work has been funded by the Swiss Innovation Agency (grant \textnumero 37896) and by the Swiss National Science Foundation under NCCR Automation (grant \textnumero 180545).}
\thanks{$^1$ Xavier Guidetti, Alisa Rupenyan, and John Lygeros are with the Automatic Control Laboratory, ETH Zürich, Switzerland {\tt\footnotesize\{xaguidetti, ralisa, lygeros\}@control.ee.ethz.ch}}
\thanks{$^2$ Xavier Guidetti and Alisa Rupenyan are also with Inspire AG, Zürich, Switzerland}
\thanks{$^3$ Lutz Fassl and Majid Nabavi are with the Equipment Digitalization Team, Oerlikon Metco, Switzerland {\tt\footnotesize\{lutz.fassl, majid.nabavi\}@oerlikon.com}}

\thanks{Digital Object Identifier (DOI): see top of this page.}

}

\markboth{IEEE Robotics and Automation Letters. Preprint Version. Accepted September, 2022}
{Guidetti \MakeLowercase{\textit{et al.}}: Advanced Manufacturing Configuration by Sample-efficient Batch Bayesian Optimization} 

\maketitle

\begin{abstract}
We propose a framework for the configuration and operation of expensive-to-evaluate advanced manufacturing methods, based on Bayesian optimization. The framework unifies a tailored acquisition function, a parallel acquisition procedure, and the integration of process information providing context to the optimization procedure. \cmtb{The novel acquisition function is demonstrated, analyzed and compared on state-of-the-art benchmarking problems. We apply the optimization approach to atmospheric plasma spraying and fused deposition modeling.} Our results demonstrate that the proposed framework can efficiently find input parameters that produce the desired outcome and minimize the process cost.
\end{abstract}



\cmt{\section{Introduction}}


\IEEEPARstart{I}{n} manufacturing, the optimization of process inputs for a new production task to meet productivity and quality requirements, is a challenging task. This is true especially for processes where the direct outcome of the process inputs comprises several interconnected outputs, whose quality analysis requires time-consuming or destructive measurements. Additive manufacturing, and in general the technologies that deposit material layer by layer on a substrate, are examples of such processes \cite{Dey_FDM}. A sample-efficient, data-driven approach to find optimal process parameters for a manufactured part is thus beneficial, especially when the number of possible trials is restricted by the produced pieces quality assessment costs.

The standard approach for process \cmtb{configuration, control and optimization uses modeling based on statistical principles \cite{ibanez, rubio}.} Candidate combinations of process parameters are proposed following a full factorial or fractional design of experiments, using optimal orthogonal designs or Taguchi arrays to limit the number of experiments \cite{Gao_2012}. Then, the best candidates are selected following regression analysis of the experimentally determined quality parameters \cite{Datta2013ModelingAnalysis, Wu2015EmpiricalProcess}. Alternatively, data-driven modeling using neural networks \cite{Kanta2008ArtificialProcesses, Kanta2009ArtificialAttributes} or support vector machines \cite{DING_waam} has been proposed to relate process inputs to quality parameters. All these approaches require a large number of samples, to either achieve good predictive capabilities, or to reliably cover all possible process variations. 

Previous work on data-driven optimization of manufacturing processes \cite{Maier2019BayesianTurning, MaierSelf-OptimizingOptimization} has shown that Bayesian optimization (BO) is an information efficient and effective technique to automate the configuration of industrial processes with a limited number of experiments. BO has been applied in parallel or nested formulations for the tuning of algorithms \cite{Snoek2012PracticalAlgorithms}, the optimization of black-box functions \cite{Wang2020ParallelFunctions, Ginsbourger2008AProcesses, Azimi2012HybridOptimization}, or the tuning of cascaded controllers \cite{konig2020safety}. Safety-aware BO methods have been used to ensure that safety constraints are respected, providing probabilistic guarantees that all candidate samples remain in the constraint set \cite{Sui}. They have been demonstrated for robotic applications \cite{Berkenkamp2} and for adaptive control in position tracking \cite{konig2021safe}. Iterative model-based learning and control methods have recently been applied to specific processes in the field of additive manufacturing \cite{RL_WAAM}.

In this paper, we propose a data-driven approach for the optimization of process input parameters, given desired output properties of the manufactured components. The main contributions of our work are:
\begin{enumerate*}
    \item a sample-efficient parameters selection procedure, based on a novel BO acquisition function whose aggressiveness can be tuned,
    \item \cmtb{a detailed analysis of the novel acquisition function performance on benchmarking problems from the literature, and}
    \item a parallelized status-aware optimization procedure that incorporates process information in the BO procedure, making it fully applicable to \cmtb{any experimental scenario}.
\end{enumerate*}
We demonstrate the proposed method on atmospheric plasma spraying (APS) \cmtb{and fused deposition modeling (FDM, also known as 3D printing), which perfectly exemplify the challenges of advanced manufacturing -- namely, hard-to-model multiple-input-multiple-output relationships, expensive deposition trials and time-consuming quality characterization of manufactured pieces.}

\cmtb{In Section \ref{sec_background}, we detail the challenge that motivated the research and the techniques upon which we base our work. Section \ref{sec_method} presents our contributions on BO. In Section \ref{sec:comparative_study} we conduct a thorough analysis and comparison of the proposed optimization method. Section \ref{sec:implementation} extends the BO method to complex manufacturing applications. Lastly, Sections \ref{sec_plasma} and \ref{sec:FDM} detail the experimental work conducted on APS and FDM.}

\section{Background} \label{sec_background}

\subsection{Process optimization in advanced manufacturing} \label{sec_problem}

Multiple advanced manufacturing processes, especially those relying on layer-wise deposition, are challenging for precise modeling and optimization, due to their inherent complexity.
Data-driven optimization is impeded by limited data availability, which is common in this class of problems. The process outputs, i.e. the desired part properties, are unknown and can only be evaluated point-wise. The properties analysis is a slow and expensive procedure, requiring destructive approaches.
The configuration is often accelerated by performing experiments in batches. Between experimental sessions, the manufacturing equipment often undergoes changes (wearing, maintenance, etc.) that need to be tracked to reduce output variation.
The production cost is represented by a deterministic function of the input parameters (e.g. used resources, consumed energy, induced equipment wear, etc.). 
In this manufacturing context, there exists a peculiar relationship existing between cost reduction and constraints fulfillment. The main focus of the optimization lies in finding \emph{feasible} samples, while cost reduction is a secondary goal to be achieved once feasibility has been reached. Furthermore, as these processes are often used to manufacture very small lots, producing as many feasible samples as possible during configuration itself is important \cite{I4_0review}. While suboptimal in terms of cost, these samples are usable, reducing the amount of runs conducted and the total configuration cost.

\subsection{Gaussian Processes} \label{sec_GP}

Given $\mathbf{x} \in \mathbb{R}^n$ where $n$ is the number of inputs, we model each output function $c(\mathbf{x})$ using Gaussian process regression. A Gaussian process (GP) is a collection of random variables, any finite number of which have a joint Gaussian distribution. It provides a distribution over functions $c(\mathbf{x}) \sim \mathcal{GP}(\mu(\mathbf{x}),k(\mathbf{x},\mathbf{x}'))$ that is fully defined by its mean function $\mu(\mathbf{x})$ and its covariance, given by the kernel function $k(\mathbf{x},\mathbf{x}')$. We denote the $i$-th measurement corresponding to an input vector $\mathbf{x}_i$ by $y_i=c(\mathbf{x}_i)+\varepsilon_i$, where $\varepsilon_i$ is the measurement noise with distribution $\mathcal{N}(0,\sigma^2_\mathrm{n})$. Given a set of $p$ input vectors paired with the corresponding noise corrupted measurements $\mathcal{T} = \{(\mathbf{x}_i,y_i)\}_{i=1}^p$, we can calculate the posterior distribution of $c(\cdot)$ at any query point $\bar{\mathbf{x}}$. Denoting the set of inputs $\mathbf{X} = \{\mathbf{x}_i\}_{i=1}^p$ and the set of corresponding measurements $\mathbf{y} = \{y_i\}_{i=1}^p$, we obtain $\Tilde{c}(\bar{\mathbf{x}}) \sim \mathcal{N}\left(\mu_c(\bar{\mathbf{x}}),\sigma^2_c(\bar{\mathbf{x}})\right)$, where the corresponding posterior mean and variance are given as
\begin{align}
\mu_c(\bar{\mathbf{x}}) &= \mu(\bar{\mathbf{x}}) + k(\bar{\mathbf{x}},\mathbf{X})[k(\mathbf{X},\mathbf{X})+\sigma^2_\mathrm{n}\mathbb{I}]^{-1}(\mathbf{y}-\mu(\mathbf{X})) \, , \nonumber \\
\sigma^2_c(\bar{\mathbf{x}}) &= k(\bar{\mathbf{x}},\bar{\mathbf{x}}) - k(\bar{\mathbf{x}},\mathbf{X})[k(\mathbf{X},\mathbf{X})+\sigma^2_\mathrm{n}\mathbb{I}]^{-1}k(\mathbf{X},\bar{\mathbf{x}}) \nonumber \, .
\end{align}

After model training, the posterior mean $\mu_c(\bar{\mathbf{x}})$ and variance $\sigma^2_c(\bar{\mathbf{x}})$ can be used respectively as the model prediction and corresponding uncertainty at point $\bar{\mathbf{x}}$. To do so, a confidence interval of 95\% is typically selected.
A complete overview of GPs and their practical use can be found in \cite{Rasmussen2006GaussianLearning}.

\cmt{\subsection{Constrained Bayesian Optimization}} \label{sec:CBO}

In its simplest form, BO is a sequential strategy for the optimization of expensive-to-evaluate functions, often subject to safety or performance constraints. BO is commonly used with GP models, which use the available evaluations to produce a probabilistic distribution of the functions and can be updated when new samples are added to the known experiment set. To find the optimal inputs $\mathbf{x}^*$ of \cmtb{a general} constrained problem
\begin{equation}\label{eqn:CBO}
\begin{array}{cl}    
\minOp_{\mathbf{x} \in \mathcal{X}}\!
&
f(\mathbf{x})
\\\text{s.t.}
&
c(\mathbf{x}) \leq \lambda\,, 
\end{array}
\end{equation}
\cmtb{where $\lambda$ is a constant and $\mathcal{X}$ a known bounded domain,} the method starts by placing priors over the unknown objective and constraint functions $f(\mathbf{x})$ and $c(\mathbf{x})$ and then updates them with the collected data to form a posterior distribution of the functions. 
The posterior distribution is then used to select the next candidate for evaluation $\mathbf{x}_{m+1}$, according to
\begin{equation}\label{eqn:x_next}
    \mathbf{x}_{m+1} = \argmax_{\mathbf{x}\in\mathcal{X}} \alpha_m(\mathbf{x}),
\end{equation}
where $\alpha_m(\mathbf{x})$ is the acquisition function built based on the $m$ previously evaluated inputs. Well-designed acquisition functions trade off exploration and exploitation by combining the information content at the inputs and the corresponding predicted performance. They are a central ingredient of BO -- and of our proposed approach -- and can be tailored to specific (classes of) optimization problems. Numerous acquisition functions have been proposed in more or less recent works \cite{Guinet2020Pareto-efficientOptimization,Hernandez-Lobato2016ASearch,Garrido-Merchan2019PredictiveConstraints}.
In constrained optimization, the acquisition function considers both the expected objective improvement and the expected feasibility of inputs, to select cost-reducing candidates that fulfill the constraints with high probability\cite{Gardner2014BayesianConstraints}.


\section{Method} \label{sec_method}

We consider the class of optimization problems having a deterministic objective and one or more black-box constraints. We write our optimization problem as
\begin{align}
\min_{\mathbf{x} \in \mathcal{X}}\quad &S(\mathbf{x}) \label{eq:opt_problem} \\ 
\text{s.t.} \quad & c_k(\mathbf{x}) \leq \lambda_k\ , \quad k = 1,\ldots,K , \nonumber
\end{align}
where $S(\mathbf{x})$ denotes the problem objective, $c_k(\mathbf{x})$ the $k^{\textrm{th}}$ constraint, and $K$ the number of constraints. The input combinations $\mathbf{x}$  belong to a known bounded domain $\mathcal{X}$. We assume that $S(\mathbf{x})$ can be freely computed as a deterministic function of inputs $\mathbf{x}$, whereas all $c_k(\mathbf{x})$ are unknown\cmtb{, can have any complexity, and are} sampled at each iteration. We also assume that the optimization can be initialized with a set of previously evaluated input vectors and corresponding constraints values $\mathcal{T}=\{\mathbf{x}_i,\mathbf{c}(\mathbf{x}_i)\}_{i=1}^p$ where, for each $i$, $\mathbf{c}(\mathbf{x}_i) = \{c_k(\mathbf{x}_i)\}_{k=1}^K$.

\subsection{Acquisition Procedure} \label{sec_acquisition_procedure}

We propose a custom acquisition procedure tailored to the problems at hand. We first introduce two functions: \textit{improvement} and \textit{feasibility probability}. The first one determines the amount of improvement, in terms of cost reduction, that a vector with a candidate combination of inputs can produce. We define it as
\begin{equation}
    I(\mathbf{x}) = \max \left\{0, S(\mathbf{x}^+)-S(\mathbf{x}) \right\} \label{eq:imp}\\,
\end{equation}
where $\mathbf{x}^+$ is the feasible combination of inputs with the lowest cost found so far. If no feasible point is known, we set $S(\mathbf{x}^+) = \max_{\mathbf{x} \in \mathcal{X}}S(\mathbf{x})+1$. Candidate combinations producing a cost higher than $\mathbf{x}^+$ return no improvement. If the cost $S(\mathbf{x})$ was unknown and not deterministic, we would need to take the expectation of \eqref{eq:imp}, corresponding to the \emph{expected improvement} acquisition function of \cite{Gardner2014BayesianConstraints}. However, here we do not need to take an expectation of \eqref{eq:imp} as, knowing $S(\mathbf{x})$, $I(\mathbf{x})$ is also deterministic.

To include the constraints of \eqref{eq:opt_problem} we define the feasibility probability as
\begin{align}
    \text{\em{FP}}(\mathbf{x}) = \text{\em{Pr}}[\Tilde{c}(\mathbf{x}) \leq \lambda] 
    = \int_{-\infty}^{\lambda} p(\Tilde{c}(\mathbf{x})|\mathbf{x},\mathcal{T})d\Tilde{c}(\mathbf{x}) \ .
\end{align}
Since $\Tilde{c}(\mathbf{x})$ has Gaussian marginals, $\text{\em{FP}}(\mathbf{x})$ is a Gaussian cumulative distribution function. With multiple independent constraints, the feasibility probability is
\begin{equation}
    \text{\em{FP}}(\mathbf{x}) = \prod_{k=1}^K \text{\em{Pr}}[\Tilde{c}_k(\mathbf{x}) \leq \lambda_k] \label{eq:fp}\ .
\end{equation}
We now define the two \cmtb{novel} acquisition functions that our algorithm exploits: 
\begin{subequations}
\begin{align}
    \alpha_{\text{\em{FIP}}}(\mathbf{x}) &= \text{\em{FP}}(\mathbf{x})\text{sgn}\{I(\mathbf{x})\} \label{fip} \\
    \alpha_{\text{\em{HFI}}}(\mathbf{x}) &= (\text{\em{FP}}(\mathbf{x})-\pi)I(\mathbf{x}) \label{hfi} \ ,
\end{align}
\end{subequations}
where $\pi \in [0, 1]$ is a confidence threshold that tunes the aggressiveness of our acquisition algorithm. \textit{Feasible Improvement Probability} (FIP) \eqref{fip} has a conservative approach: it returns the feasibility probability of the candidates that are known to produce any cost improvement. As the magnitude of the improvement is eliminated by the sign function, maximizing $\alpha_{\text{\em{FIP}}}(\mathbf{x})$ corresponds to looking for the points that have the highest chances of respecting the constraints. \textit{High FIP Improvement} (HFI) \eqref{hfi} is more aggressive: the magnitude of the cost improvement modulates the candidate selection, producing a trade-off between feasibility probability and reward. Maximizing $\alpha_{\text{\em{HFI}}}(\mathbf{x})$ returns candidates that markedly reduce the cost while maintaining a minimum feasibility probability of $\pi$.

\vspace{-3mm}
\begin{algorithm}[htbp]
\footnotesize

\SetAlgoLined
\LinesNumbered
\SetKwInOut{Input}{input}
\Input{$\text{\em{FP}}(\mathbf{x})$ and $I(\mathbf{x})$ of all candidates $\mathbf{x}$ in the candidates set $\mathcal{U}$, previously evaluated inputs set $\mathcal{T}$, constraints, threshold probability $\pi$}
Compute $\alpha_{\text{\em{FIP}}}(\mathbf{x})$ and $\alpha_{\text{\em{HFI}}}(\mathbf{x})$ for all candidates\;
Group the elements of $\mathcal{T}$ respecting the constraints in $\mathcal{T}_\mathrm{f} \subset \mathcal{T}$\;
\eIf{$\mathcal{T}_\mathrm{f} = \emptyset$}{
$\alpha(\mathbf{x}) \longleftarrow \alpha_{\text{\em{FIP}}}(\mathbf{x})$ for all $\mathbf{x} \in \mathcal{U}$ \label{line_nofeas}\;
}{
\eIf{any candidate verifies $\alpha_{\text{\em{FIP}}}(\mathbf{x}) > \pi$ \label{line_pi}}{
$\alpha(\mathbf{x}) \longleftarrow \alpha_{\text{\em{HFI}}}(\mathbf{x})$ for all $\mathbf{x} \in \mathcal{U}$\;}{
$\alpha(\mathbf{x}) \longleftarrow \alpha_{\text{\em{FIP}}}(\mathbf{x})$ for all $\mathbf{x} \in \mathcal{U}$ \label{line_low_feas}\;}
}
\Return{selected candidate $\mathbf{x}^* = \argmax_{\mathbf{x} \in \mathcal{U}} \alpha(\mathbf{x})$ \label{line_maxim}}
 \caption{Candidate Selection}
 \label{alg_candsel}
\end{algorithm}
\vspace{-3mm}

Algorithm \ref{alg_candsel} presents the complete candidate selection procedure for a candidate set $\mathcal{U}$. As the results of Sections  \ref{sec:pi_study} and \ref{sec_simulated_res} will show, we observed that \cmtb{state-of-the-art methods} such as \cite{Gardner2014BayesianConstraints} are too aggressive, while an approach focusing on FIP only led to excessively conservative exploration. Given that maintaining feasible solutions has higher priority than optimizing $S(\mathbf{x})$, we \cmtb{introduce a novel switching} acquisition procedure that maintains a trade-off between the two approaches. As long as no feasible point is found, we exclusively focus on maximizing the chances of finding one. We therefore perform the optimization according to \eqref{fip} (line \ref{line_nofeas}). Once we have found a feasible experiment, we take a mixed approach. We want to ensure that the aggressive exploration of \eqref{hfi} is conducted only with a sufficient safety margin, given by the confidence threshold $\pi$ (line \ref{line_pi}). When the probability of finding cost reducing points is too low, we take the conservative approach of \eqref{fip} (line \ref{line_low_feas}). At line \ref{line_maxim}, we select the candidate belonging to the candidates set $\mathcal{U}$ that has been assigned the largest $\alpha(\mathbf{x})$. This procedure increases the amount of feasible experiments found during the optimization and favors a safer exploration of the constraints space over uncertain large improvements. It is important to remark that, given the deterministic nature of our objective function, both \eqref{fip} and \eqref{hfi} only consider candidates that will certainly produce a cost reduction.


\cmt{\section{Acquisition Procedure Analysis} \label{sec:comparative_study} 

In this section, we evaluate the performance of the acquisition function proposed in Section \ref{sec_acquisition_procedure} on benchmark problems, using a Monte Carlo approach. Specifically, we compare the performance of the acquisition functions proposed in Alg. \ref{alg_candsel} with that of the expected constrained improvement ($EI_C$) \cite{Gardner2014BayesianConstraints} acquisition function, which has been successfully applied to similar industrial application and represents the state of the art in the domain of manufacturing processes configuration via Bayesian optimization. We selected three 2D problems that have been repeatedly used as benchmarks in the literature:

\vspace{-3mm}
{\footnotesize
\begin{align}
        \minOp_{\mathbf{x} \in [0,6]^2}\quad f(\mathbf{x}) &= \cos{(2x_1)}\cos{(x_2)} + \sin{(x_1)} \nonumber \tag{P1} \label{eq:P1}\\ 
        \text{s.t.} \quad  c(\mathbf{x}) &= \cos{(x_1)}\cos{(x_2)}-\sin{(x_1)}\sin{(x_2)} \leq -0.5 \,; \nonumber  \\[2ex]
        \minOp_{\mathbf{x} \in [0,6]^2}\quad f(\mathbf{x}) &= \sin{(x_1)} + x_2 \nonumber  \tag{P2} \label{eq:P2} \\ 
        \text{s.t.} \quad  c(\mathbf{x}) &= \sin{(x_1)}\sin{(x_2)} \leq -0.95 \,; \nonumber \\[2ex] 
        \min_{\mathbf{x} \in [0,1]^2}\quad f(\mathbf{x}) &= x_1 + x_2 \nonumber  \tag{P3} \label{eq:P3} \\
        \text{s.t.} \quad  c_1(\mathbf{x}) &= \frac{3}{2}-x_1-2x_2-\frac{1}{2}\sin{(2\pi(x_1^2-2x_2))} \leq 0 \nonumber  \\
         c_2(\mathbf{x}) &= x_1^2 + x_2^2 - \frac{3}{2}  \leq 0 \,. \nonumber
\end{align}
}

Problems \ref{eq:P1} and \ref{eq:P2} originally appeared in \cite{Gardner2014BayesianConstraints}, while Problem \ref{eq:P3} comes from \cite{Gramacy2016ModelingOptimization}. All of them have been reused in later works, such as \cite{Letham2017ConstrainedExperiments}. We consider the objectives $f(\mathbf{x})$ to be known and the constraints $c(\mathbf{x})$ to be unknown, and learned by sampling during the BO, or in an offline training phase. Both problems \ref{eq:P1} and \ref{eq:P2} have complicated objective and constraint functions with two disjoint feasible regions. The feasible domain of \ref{eq:P2} has a smaller total surface than the one of \ref{eq:P1}, making it harder to find feasible samples in \ref{eq:P2}. The objective function in \ref{eq:P3} is simple, but its two constraint functions form a complicated feasible region, which makes \ref{eq:P3} the most interesting problem in the context of this work. We conducted the study both for noiseless and noisy scenarios.

\subsection{Noiseless Scenario}\label{sec:noisless_study}

In the noiseless case, where the constraints evaluations are not corrupted, we set $\pi=0.6$, corresponding to a moderately aggressive search for the optimal parameters. We grid each problem's input space to generate 20000 candidates. The optimization is stopped when the best candidate is evaluated, or when 100 iterations have been conducted, and the corresponding iteration number is called \textit{required iterations}. Evaluated inputs $\mathbf{x}$ respecting the problem constraints are called \textit{feasible samples}. We begin each optimization procedure by randomly selecting two initialization samples. These are provided to both Alg. \ref{alg_candsel} and $EI_C$. Each optimization procedure is repeated 100 times with random initialization samples. Figure \ref{fig:toy_comparisons} shows the benchmark problems together with one representative optimization trace.
\begin{figure}[htbp]
     \centering
     \begin{subfigure}[b]{\columnwidth}
        \begin{minipage}{.52\textwidth}
            \centering
            \includegraphics{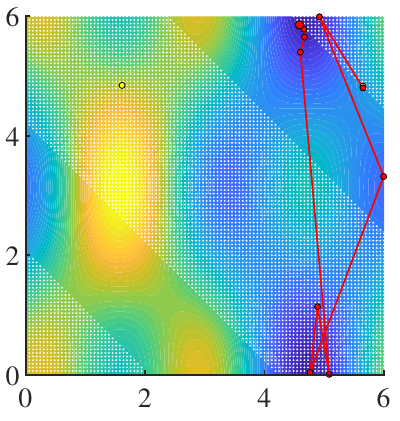}
        \end{minipage}%
        \begin{minipage}{.48\textwidth}
            \centering
            \includegraphics{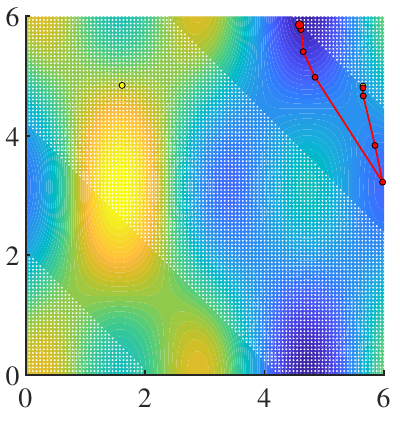}
        \end{minipage}
        \vspace{-2mm}
         \caption{Problem \ref{eq:P1}}
     \end{subfigure}
     \hfill
     \begin{subfigure}[b]{\columnwidth}
         \begin{minipage}{.52\textwidth}
            \centering
            \includegraphics{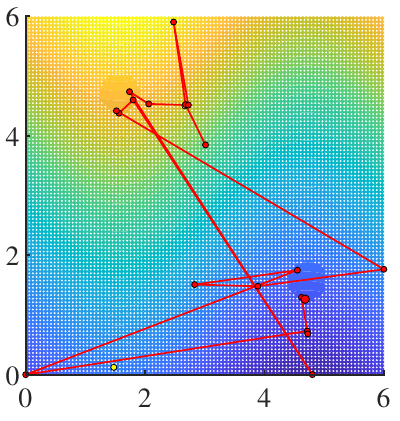}
        \end{minipage}%
        \begin{minipage}{.48\textwidth}
            \centering
            \includegraphics{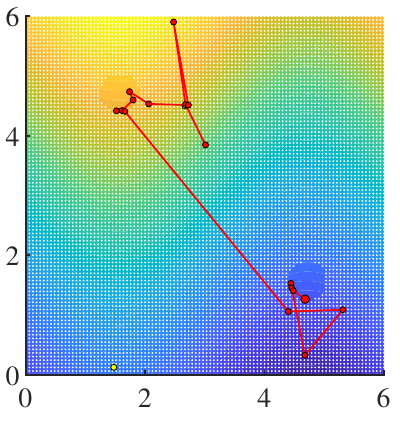}
        \end{minipage}
        \vspace{-2mm}
         \caption{Problem \ref{eq:P2}}
     \end{subfigure}
     \hfill
     \begin{subfigure}[b]{\columnwidth}
         \begin{minipage}{.5\textwidth}
            \centering
            \includegraphics{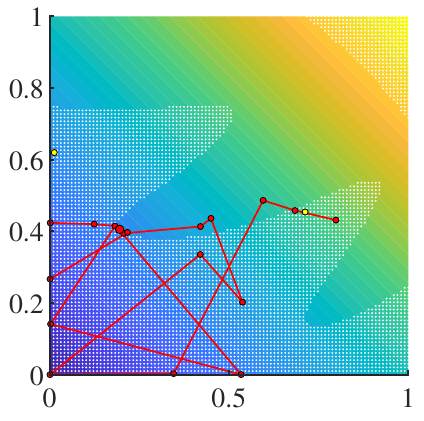}
        \end{minipage}%
        \begin{minipage}{.5\textwidth}
            \centering
            \includegraphics{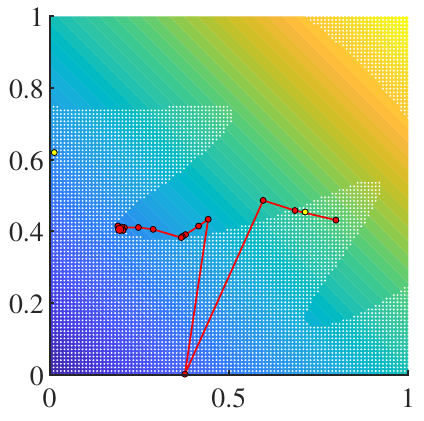}
        \end{minipage}
        \vspace{-2mm}
         \caption{Problem \ref{eq:P3}}
     \end{subfigure}
        \caption{Comparison of representative optimization traces for $EI_C$ (left) and Alg. \ref{alg_candsel} (right). The problems objective is represented with a colormap, with lower values plotted in blue. White dots are used to show unfeasible regions. Initialization samples are marked in yellow, while the trace is shown in red, with a larger final marker.}
        \label{fig:toy_comparisons}
\end{figure}
The mixed strategy upon which Alg. \ref{alg_candsel} is based results in a sequence of evaluations that generally avoid large constraint violations, and produces a larger fraction of \textit{feasible samples} than $EI_C$. It can be seen in all cases that $EI_C$ selects samples that would produce very large cost reductions, but turn out to be infeasible, effectively wasting evaluations. Because of this, Alg. \ref{alg_candsel} often requires a comparable or smaller number of evaluations to find the optimizer. In the situations where $\alpha_{\text{\em{FIP}}}(\mathbf{x}) > \pi$, Alg. \ref{alg_candsel} has been designed to behave very similarly to $EI_C$, which explains why some sections of the optimization traces look identical for both approaches.

\begin{figure}[htbp]
     \centering
     \begin{subfigure}[b]{\columnwidth}
         \centering
         \includegraphics{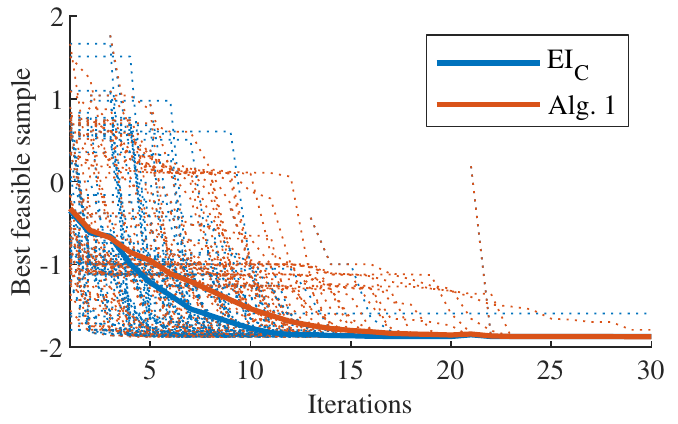}
         \vspace{-2mm}
         \caption{Problem \ref{eq:P1}}
         \label{subfig:convergence_p1}
     \end{subfigure}
     \begin{subfigure}[b]{\columnwidth}
         \centering
         \includegraphics{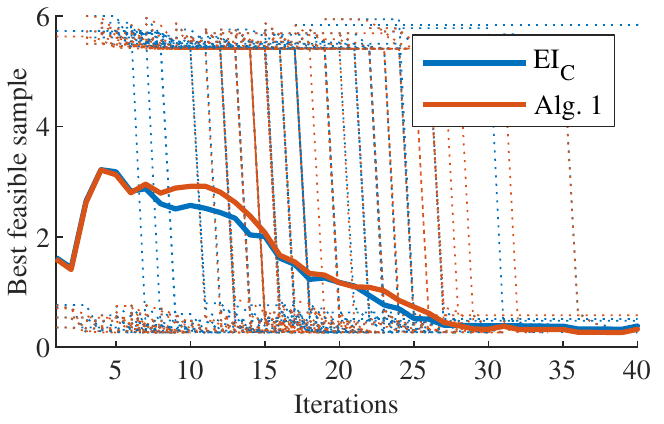}
         \vspace{-2mm}
         \caption{Problem \ref{eq:P2}}
     \end{subfigure}
     \begin{subfigure}[b]{\columnwidth}
         \centering
         \includegraphics{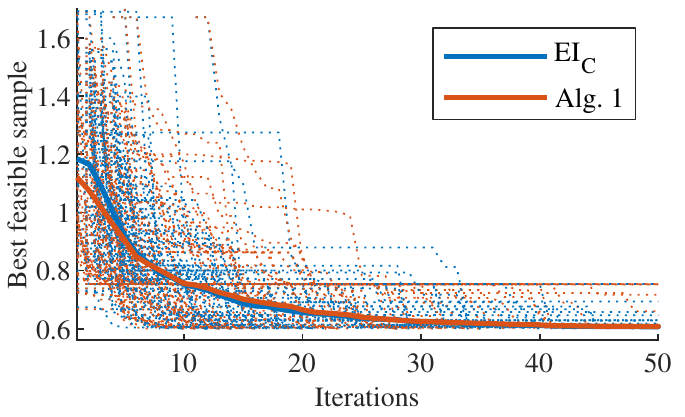}
         \vspace{-2mm}
         \caption{Problem \ref{eq:P3}}
     \end{subfigure}
        \caption{Convergence speed comparison over 100 repetitions. Dashed lines represent best feasible sample found at a given iteration for each repetition. These results are averaged to produce the solid lines.}
        \label{fig:convergence}
\end{figure}
Fig. \ref{fig:convergence} shows that both algorithms converge with similar rates. As seen in Fig. \ref{fig:toy_comparisons}, $EI_C$ aggressiveness often makes feasible samples harder to find. $EI_C$ shows a marginally better convergence speed only for the first problem, as can be seen on Fig. \ref{subfig:convergence_p1}. This is due to the relatively simple constraint of Problem \ref{eq:P1} that makes the cautiousness of Alg. \ref{alg_candsel}  unnecessary.

The comparison of achieved averaged feasible samples and required iterations across all 100 repetitions is summarized in Table \ref{tab_noiseless_comp}. In all three problems, Alg. \ref{alg_candsel} outperforms $EI_C$ by producing more \textit{feasible samples}, without an adverse effect on the number of \textit{required iterations}. Increasingly complex constraints tend to favor Alg. \ref{alg_candsel}, that outperforms $EI_C$ both in terms of \textit{required iterations} and in terms of \textit{feasible samples} for Problems \ref{eq:P2} and \ref{eq:P3}.
\begin{table}[htbp]
\caption{Noiseless Scenario Comparison}
\centering
\renewcommand{\arraystretch}{1}
\begin{tabular}[h]{@{}l r r c r r c r r @{}}\toprule
& \multicolumn{2}{c}{\ref{eq:P1}}  && \multicolumn{2}{c}{\ref{eq:P2}} && \multicolumn{2}{c}{\ref{eq:P3}}\\
\cmidrule{2-3} \cmidrule{5-6} \cmidrule{8-9}
&  Alg. \ref{alg_candsel} & $EI_C$ && Alg. \ref{alg_candsel} & $EI_C$ && Alg. \ref{alg_candsel} & $EI_C$ \\
 \midrule
Req. it.   & 14.6 & 13.0 && 22.2 & 22.9 && 26.1 & 31.8\\
Feas. sam.      & 66\% & 49\% && 35\% & 28\% && 54\% & 27\%\\
\bottomrule
\label{tab_noiseless_comp}
\end{tabular}
\vspace{-8mm}
\end{table} 

\subsection{Noisy Scenario}

In this comparison, we corrupt each constraint evaluation by adding normally distributed noise $\mathcal{N}(0,\tau^2)$, using $\tau = 0.2$, following \cite{Letham2017ConstrainedExperiments}. As in the previous study, we keep $\pi=0.6$. We utilize the same list of 20000 candidates that was generated for the noiseless case. We stop the optimization when a feasible candidate in a tolerance radius (0.15 for \ref{eq:P1} and \ref{eq:P2}, and 0.0125 for Problem \ref{eq:P3}) from the best candidate is evaluated, or after conducting 100 iterations. We consider as feasible samples the evaluated inputs $\mathbf{x}$ whose corrupted evaluation respects the constraints. As in the noiseless case, we initialize Alg. \ref{alg_candsel} and $EI_C$ with two identical random samples. Each algorithm is run five times with the same initialization samples, but different noise realizations. We repeat this with 20 different initializations for a total of 100 optimizations. The aggregated results, averaged across all repetitions, are shown in Table \ref{tab_noisy_comp}.
\begin{table}[htbp]
\caption{Noisy Scenario Comparison}
\centering
\renewcommand{\arraystretch}{1}
\begin{tabular}[h]{@{}l r r c r r c r r @{}}\toprule
& \multicolumn{2}{c}{\ref{eq:P1}}  && \multicolumn{2}{c}{\ref{eq:P2}} && \multicolumn{2}{c}{\ref{eq:P3}}\\
\cmidrule{2-3} \cmidrule{5-6} \cmidrule{8-9}
&  Alg. \ref{alg_candsel} & $EI_C$ && Alg. \ref{alg_candsel} & $EI_C$ && Alg. \ref{alg_candsel} & $EI_C$ \\
 \midrule
Req. it.   & 26.0 & 28.5 && 60.9 & 49.9 && 32.5 & 36.0\\
Feas. sam.      & 59\% & 35\% && 25\% & 15\% && 48\% & 19\%\\
\bottomrule
\label{tab_noisy_comp}
\end{tabular}
\vspace{-3 mm}
\end{table} 
As expected, the performance of both algorithms deteriorates compared to the noiseless case for all problems. Despite this, Alg. \ref{alg_candsel} still outperforms $EI_C$ in terms of feasible samples for all Problems, and the number of required iterations is again comparable. Only in the case of Problem \ref{eq:P2}, Alg. \ref{alg_candsel} requires on average 11 more iterations than $EI_C$ to meet the stopping condition. This is explained by the noise magnitude, which is large when compared to the feasible region size. Furthermore, when Alg. \ref{alg_candsel} finds the top-left feasible region, its cautious nature makes it less likely to explore the bottom-right feasible region where the optimizer is. The difference between the two algorithms is particularly visible in Problem \ref{eq:P3}, that has the most complex constraints. Alg. \ref{alg_candsel} finds 2.5 times more feasible samples than $EI_C$, and the efficient learning of the feasible region leads to faster convergence.

\subsection{Confidence Threshold Study} \label{sec:pi_study}

We now analyze the effect of the confidence threshold $\pi$ on Alg. \ref{alg_candsel} performance, focusing on Problem \ref{eq:P3}. Its simple objective and complex constraints exemplify the class of problems for which Alg. \ref{alg_candsel} was designed. We follow the same procedure as in Sec. \ref{sec:noisless_study}. We repeat the numerical study with 11 different values of $\pi$ to produce the data shown in Fig. \ref{fig:pi_study}.

\begin{figure}[htbp]
\centerline{\includegraphics{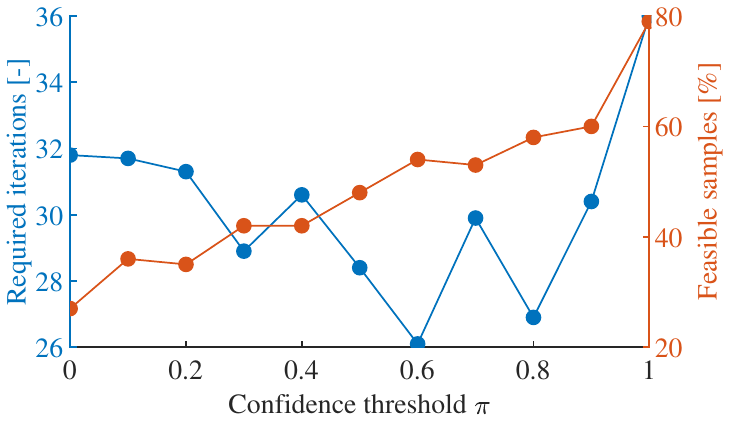}}
\caption{\textit{Required iterations} and \textit{feasible samples} for different values of $\pi$, when solving Problem \ref{eq:P3} using Alg. \ref{alg_candsel}.}
\label{fig:pi_study}
\end{figure}

The confidence threshold $\pi$ modulates the behavior of our algorithm. When $\pi = 0$, Alg. \ref{alg_candsel} only uses \eqref{hfi}, which in this case is identical to using $EI_C$ on a known objective function. As expected, the results for $\pi = 0$ correspond to the ones in Table \ref{tab_noiseless_comp} for $EI_C$ on \ref{eq:P3}. At the other extreme, when $\pi = 1$, Alg. \ref{alg_candsel} only uses \eqref{fip}, which produces a very conservative and slow approach that never attempts bold cost reductions. We further observe that increasing the value of $\pi$ increases the fraction of \textit{feasible samples} in the optimization trace. Interestingly, the number of \textit{required iterations} does not grow with $\pi$. Instead, given the nature of the problem, a cautious learning of the constraints can accelerate the optimization. All values of $\pi \in [0.1,0.9]$ make Alg. \ref{alg_candsel} outperform $EI_C$ on both metrics, with a minimum number of required iterations found at $\pi=0.6$. While the confidence threshold can be fine-tuned according to the optimization goals, the algorithm is relatively robust to the changes of $\pi$ and can be safely set to $\pi \in [0.3,0.8]$ to produce satisfactory results. Tuning-dependent algorithms often have a very narrow hyper-parameter window, out of which performance deteriorates quickly. This is not the case with Alg. \ref{alg_candsel}: the robust performance and its gradual change make the fine-tuning task simpler for practitioners.} \cmtb{The complexity and computation time of Alg. \ref{alg_candsel} are independant of $\pi$ and entirely similar to those of comparable methods such as $EI_C$, with each iteration being completed in less than \SI{500}{\milli\second}.}




\section{Implementation on Manufacturing Processes} \label{sec:implementation}

In this section we describe the methods we developed to be able to efficiently utilize Alg. \ref{alg_candsel} for the parameters tuning of manufacturing processes.

\subsection{Parallel Optimization} \label{sec:par_opt}

Parallelizing the BO evaluations accelerates the data collection procedures. Several methods to parallelize BO have been proposed \cite{Snoek2012PracticalAlgorithms, Wang2020ParallelFunctions, Ginsbourger2008AProcesses, Azimi2012HybridOptimization}. We utilize a simple fixed batch size technique that is largely based on sequential selection of query points. Unlike standard BO where the candidate $\mathbf{x}^*$ evaluation is conducted immediately, a prediction $\hat{y}^*$ of the output produced by the candidate is made. The data set of known evaluations is virtually expanded using the prediction, the GP is retrained, and another candidate is selected. This process of virtual evaluations is repeated until a batch $\mathcal{B} = \{\mathbf{x}^*_i|\mathbf{x}^*_i \in \mathcal{X}\}^n_{i=1}$ having the desired size $n$ has been generated. The points belonging to the batch are evaluated simultaneously and the predictions $\{\hat{y}^*_i\}_{i=1}^n$ are replaced with the experimental results $\{y^*_i\}_{i=1}^n$ to retrain the GP. When using GPs to model the unknown functions, data set virtual augmentation can be carried on by drawing samples $\hat{y}^*$ from the posterior distribution evaluated at the candidate site $\Tilde{c}(\mathbf{x}^*) \sim \mathcal{N}\left(\mu_c(\mathbf{x}^*),\sigma^2_c(\mathbf{x}^*)\right)$.
\subsection{Status-Aware Optimization}
\label{sec_opt_workflow}

Industrial process modeling can include uncontrollable but measurable parameters, reflecting the status of the manufacturing equipment. Output data collected in different sessions might contain offsets due to drifts or undocumented changes to the equipment between sessions. Including all measurable process parameters in the modeling and optimization makes it possible to prevent deviations during the configuration.

Let us consider a process whose model takes inputs $\mathbf{x} = (\mathbf{x}_\mathrm{c},\mathbf{x}_\mathrm{m})$ to predict the outputs $\mathbf{c}(\mathbf{x})$. The inputs in $\mathbf{x}_\mathrm{c}$ can be freely tuned during the BO configuration process, while $\mathbf{x}_\mathrm{m}$ represents measurements obtained at the end of each evaluation of $\mathbf{x}$. The measurements $\mathbf{x}_{\mathrm{m}}$ may depend on the controllable parameters $\mathbf{x}_\mathrm{c}$ as well as on the equipment status. We assume that this status remains constant during a single experimental session, but can change between sessions.

We detail a calibration procedure for the case where $\mathbf{x}_{\mathrm{m}}$ contains a single status dependent measurement $V$. Using the entries of the initialization data set $\mathcal{T}$, we train a model accepting the controllable inputs $\mathbf{x}_\mathrm{c}$ and predicting the measurement $\hat{V} = \mathrm{M_V} (\mathbf{x}_\mathrm{c})$. At the beginning of each new experimental session, we conduct an experiment with any settings $\mathbf{x}_\mathrm{c}^\mathrm{b} \in \mathcal{T}$, measure the corresponding $V^\mathrm{b}$ and compute the offset $\delta^\mathrm{b} = V^\mathrm{b} - \hat{V}^\mathrm{b}$, where $\hat{V}^\mathrm{b} = \mathrm{M_V} (\mathbf{x}_\mathrm{c}^\mathrm{b})$ is the measurement that an unchanged equipment -- with respect to $\mathcal{T}$ -- would produce. To generate the list of candidates for the BO session, we first grid the space of controllable inputs $\mathcal{X}_\mathrm{c}$ to produce a set of candidates $\mathbf{x}_\mathrm{c}$.
We then predict the status dependent measurement $\hat{V}_\delta$ corresponding to each candidate $\mathbf{x}_\mathrm{c}$ in the set according to $\hat{V}_\delta = \mathrm{M_T}(\mathbf{x}_\mathrm{c})+\delta^\mathrm{b}$. Each prediction is used to expand the corresponding $\mathbf{x}_\mathrm{c}$, producing a candidates set $\mathcal{U}$ of vectors $\mathbf{x} = (\mathbf{x}_\mathrm{c},\mathbf{x}_\mathrm{m})$.

\begin{algorithm}[htbp]

\footnotesize

\SetAlgoLined
\LinesNumbered
\SetKwInOut{Input}{input}
\Input{Initialization data set $\mathcal{T}$, candidate set $\mathcal{U}$, constraints $\lambda$, batch size $n$, threshold probability $\pi$, termination threshold $\epsilon$}
Create an empty candidate batch $\mathcal{B} \leftarrow \emptyset$\;
\Repeat{$\alpha_{\text{\em{FIP}}}(\mathbf{x}^*_i) < \epsilon$ for at least half of the candidates $\mathbf{x}^*_i$ selected for a batch $\mathcal{B}$ \label{alg_line_termination}}{
\If{$\mathcal{B} \neq \emptyset$}{
Evaluate experimentally the candidates in $\mathcal{B}$ and expand $\mathcal{T}$ with $\{(\mathbf{x}_i^*,\mathbf{y}_i^*)\}_{i=1}^n$, where $\mathbf{y}_i^*$ collects the evaluations of each $\mathbf{x}_i^*$\;
Empty $\mathcal{B} \leftarrow \emptyset$\;
} 
Use \eqref{eq:imp} to calculate $I(\mathbf{x})$ of all $\mathbf{x} \in \mathcal{U}$, make a virtual copy $\mathcal{T}_\mathrm{v} \leftarrow \mathcal{T}$\;
    \For{$i=1$ \KwTo $n$}{
    Using the the data in $\mathcal{T}_\mathrm{v}$, model the constraints $\mathbf{c}(\cdot)$\;
    Use \eqref{eq:fp} to calculate $\text{\em{FP}}(\mathbf{x})$ of all $\mathbf{x} \in \mathcal{U}$\;
    Select the candidate $\mathbf{x}^*_i$ using Alg. \ref{alg_candsel}\;
    Remove $\mathbf{x}^*_i$ from $\mathcal{U}$ and add it to $\mathcal{B}$\;
    Expand $\mathcal{T}_\mathrm{v}$ with $(\mathbf{x}^*_i,\mu_\mathbf{c}(\mathbf{x}^*_i))$, where $\mu_\mathbf{c}(\mathbf{x}^*_i)$ collects the predictive means of the constraints $\mathbf{c}(\cdot)$\label{alg_line_expand}\;
    $i \leftarrow i+1$\;
    }
}

\Return{feasible stress index minimizer $\mathbf{x}^+ = \argmin_{\mathbf{x} \in \mathcal{T}_\mathrm{f}} S(\mathbf{x})$, where $\mathcal{T}_\mathrm{f} \subset \mathcal{T}$ contains the feasible elements of $\mathcal{T}$}
\caption{Optimization Workflow}
\label{alg_workflow}
\end{algorithm}

\begin{figure}[htbp]
\centerline{\includegraphics{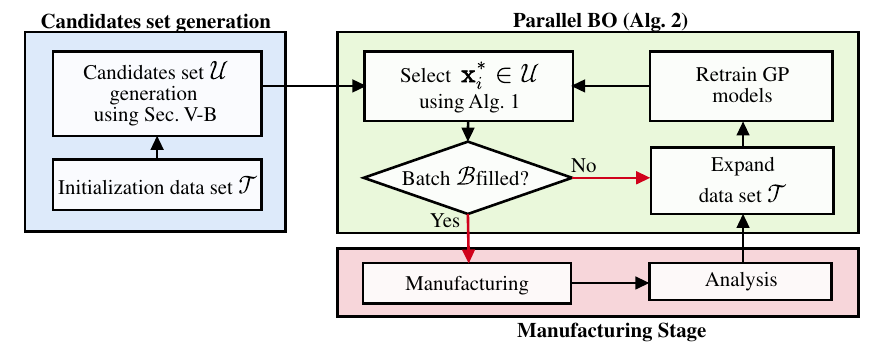}}
\caption{\cmtb{Flowchart of the proposed configuration method}}
\label{fig:flowchart}
\vspace{2mm}
\end{figure}

Parallel BO is carried on the candidate set $\mathcal{U}$ as \cmtb{detailed in Alg. \ref{alg_workflow} and Fig. \ref{fig:flowchart}}. Following the procedure described in section \ref{sec:par_opt}, we select candidates individually and expand our virtual database using the GPs posterior mean (line \ref{alg_line_expand}). The termination condition at line \ref{alg_line_termination} interrupts the procedure when most of the candidates have a low FIP, indicating that further improvement is unlikely.

\section{Atmospheric Plasma Spraying Configuration} \label{sec_plasma}

We demonstrate an application of the proposed optimization algorithm and implementation procedure on APS. APS is a thermal spraying process where micrometer-sized powder particles are injected into a viscous enthalpy plasma jet that heats them and propels them. The particles form a protective coating that improves the mechanical properties of a substrate upon bonding with its surface. The coating properties (e.g. application rate, thickness, porosity, microhardness) depend on multiple process input parameters \cite{Leblanc2000StudySpraying}.

Details about the process data-driven modeling are available in \cite{Guidetti_Plasma_BO}. Our goal is to select values for six controllable process inputs to regulate the coating microhardness and porosity and maximizing the equipment lifetime. In the absence of real-time measurements reflecting the system wear, the objective of maximizing the lifetime is encoded through the minimization of the stress index, an empirical relation reflecting the working conditions of the gun components during the coating process. The stress index value can be calculated explicitly from a subset the input parameters.
While spraying a coating, we measure the gun voltage as it contains valuable information about the equipment status.

\subsection{Simulated Process Optimization}\label{sec_simulated_res}

To conduct simulated studies, we use the \cmtb{neural network model structure and data set from \cite{Guidetti_Plasma_BO}}. The neural network simulates the behavior of the APS machine and acts as an oracle during the optimization process, returning the microhardness and porosity of virtual coated samples. 
Following the procedure detailed in section \ref{sec_opt_workflow}, we treat the gun voltage as a status dependent measurement. We simulate a scenario where, during a first gun ignition, we measure a voltage offset of \SI{2}{V}, indicating a change in the equipment status. The goal of the optimization is to find combinations of inputs producing coatings with microhardness ranging between \SI{635}{HV} and \SI{675}{HV} and porosity between 6\% and 8.2\% while minimizing the gun stress index. We select a batch size $n=5$, a threshold probability $\pi = 0.4$ and a termination threshold $\epsilon = 0.05$. We initialize the optimization with $N_{\mathrm{init}}=86$ experiments, none of which respects the constraints.

\begin{table}[htbp]
\centering
\caption{Stopping Batch and Fraction of Feasible Samples for Alg. \ref{alg_candsel} and $EI_C$}
\renewcommand{\arraystretch}{1}
\begin{tabular}[h]{@{}l r r r r @{}}\toprule
&  \makecell[c]{Alg. \ref{alg_candsel}($\pi=0.4$)} & $EI_C$\\
 \midrule
Cost & 102 & 102 \\
Stopping batch & 4 & 7\\
Nr. evaluations & 20 & 35 \\
Feasible samples & 45\% & 20\% \\
\bottomrule
\label{tab_comparison}
\end{tabular}
\vspace{-3 mm}
\end{table}

Table \ref{tab_comparison} compares the performance of the proposed acquisition procedure with that of $EI_C$ under identical conditions. Both algorithms reach the same minimum cost when meeting the termination condition, however Alg. \ref{alg_candsel} does so using 43\% less evaluations and producing 2.25 times more feasible samples. The results reaffirm those in section \ref{sec:comparative_study}, confirming that Alg. \ref{alg_candsel} outperforms $EI_C$ for APS configuration. A very detailed analysis of Alg. \ref{alg_candsel} behavior on this simulated problem is available in \cite{Guidetti_Plasma_BO}.

\cmt{\subsection{Process Optimization Experiments}} \label{sec_exp_opt}

\begin{figure}[htbp]
\centerline{\includegraphics{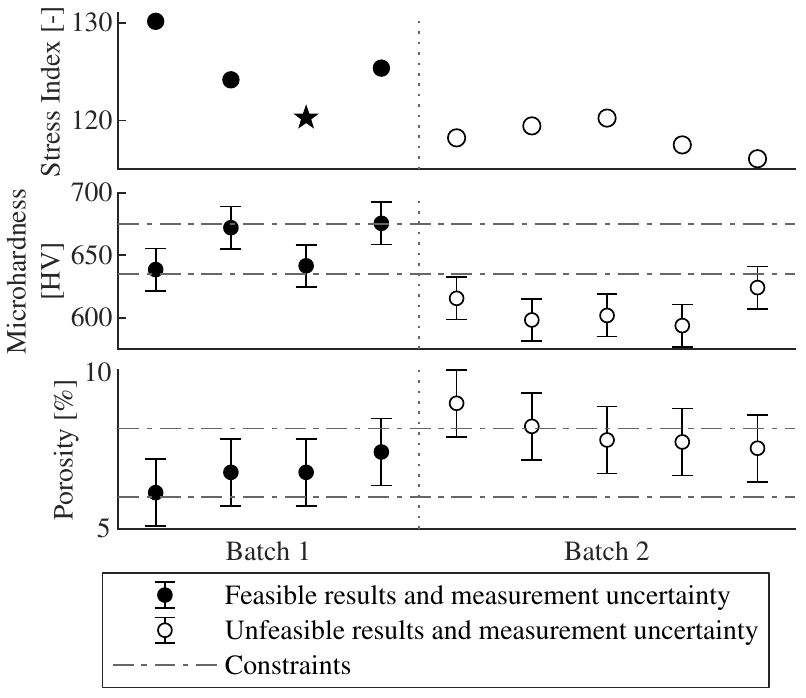}}
\caption{\cmt{Experiments for the optimization of an APS process showing the coating properties of the samples belonging to two batches. In the top panel, the star denotes the best feasible point found in the experiments.}
}
\vspace{3mm}
\label{fig_experimental_opt}
\end{figure}

We tested our algorithm on the APS machine to evaluate its real-world performance. As in the simulated case, the goal was to find combinations of inputs that produce coatings with microhardness ranging between \SI{635}{HV} and \SI{675}{HV} and porosity between 6\% and 8.2\%. The models were initialized using all the available data set of 86 experiments. We could coat four samples in Batch 1 (with a voltage offset of \SI{2}{V}) and five samples in Batch 2 (with a voltage offset of \SI{-0.8}{V}). The experimental results are shown in Fig. \ref{fig_experimental_opt}. As in the case of the simulated procedure, no initialization experiment was feasible, which made the optimization algorithm begin with a cautious approach. The samples of Batch 1 respect the imposed constraints both for microhardness and porosity. The lowest found stress index corresponds to 120.3 (indicated by a star in the figure). This value is used as an upper bound in the second batch search, where the algorithm now acts aggressively to further reduce the stress index. The results are very similar to the simulated ones, with the samples of batch 2 being unfeasible because of low microhardness (cf. \cite{Guidetti_Plasma_BO}, Fig. 5). While further experimental work was impossible \cmtb{as APS is very expensive and sample analysis extremely time-consuming}, we observe a qualitative agreement between experimental and simulated results.

\vspace{6mm}
\cmtb{
\section{Fused Deposition Modeling Configuration} \label{sec:FDM}

We further used Alg. \ref{alg_candsel} to search for optimal print parameters in FDM. We printed using a liquid-crystal polymer (LCP) filament, for which good print parameters are particularly hard to find \cite{Gantenbein}. The goal of the configuration consisted in finding a combination of extrusion rate and printer speed that would minimize the print time while maintaining a satisfactory print quality. The print quality constraint was enforced by setting an upper limit of \SI{10}{\micro\metre} to the surface roughness $R_a$ of the printed samples. We set the batch size to one and initialized Alg. \ref{alg_candsel} with seven available experiments. 

Figure \ref{fig:fdm_res} shows the results of the configuration procedure. To better demonstrate the behavior of Alg. \ref{alg_candsel}, we began the process with $\pi = 0.4$, making the approach relatively cautious. The algorithm steadily reduced the print time and a large fraction of the samples printed in this phase respect the roughness constraint. Then, after experiment 16, we lowered the confidence treshold to $\pi = 0.1$, making the algorithm more aggressive. Almost immediately, the print time was significantly reduced. As expected in this second phase, however, the fraction of samples respecting the constraint diminished significantly, clearly showing the trade-off induced by $\pi$.

\begin{figure}[htbp]
\centerline{\includegraphics{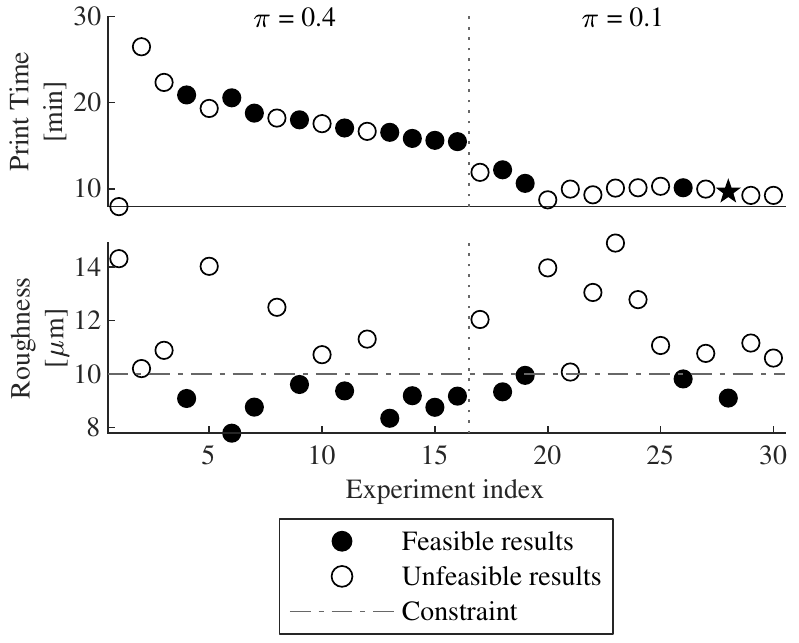}}
\caption{\cmtb{Experiments for the optimization of FDM using LCP filament, conducted with two different values of $\pi$. In the top panel, the star denotes the best feasible point found in the experiments.}}
\label{fig:fdm_res}
\end{figure}

}

\section{Conclusion}

We \cmtb{presented} a method for the automated configuration of advanced manufacturing processes, based on GP models and parallelized constrained BO. Our method incorporates process information in the optimization procedure, to efficiently direct the search for input parameters that produce the desired output property specifications and minimize the process cost. The algorithm is based on a novel acquisition method tailored to the class of problems having known objectives and black-box constraints, to which advanced manufacturing processes belong. The acquisition method performance \cmtb{was} compared to the state-of-the-art on benchmark problems. We also \cmtb{demonstrated our method on APS and FDM.} The results show that the proposed method quickly finds feasible input combinations and then exploits the collected information and the problem structure \cmtb{to optimize the processes.}

\bibliography{references_updated}
\bibliographystyle{ieeetr}

\end{document}